\documentclass[10pt,twocolumn,letterpaper]{article}

\usepackage{iccv}              
\usepackage[utf8]{inputenc} 
\usepackage[T1]{fontenc}    
 \usepackage[symbol]{footmisc}
\usepackage{url}            
\usepackage{booktabs}       
\usepackage{amsfonts}       
\usepackage{nicefrac}       
\usepackage{microtype}      
\usepackage{xcolor}         
\usepackage{makecell}

\usepackage{multirow}
\usepackage{graphicx}

\usepackage{algorithmic}
\usepackage{algorithm}

\usepackage{amsmath}
\usepackage{wrapfig}
\usepackage{subcaption} 
\usepackage{array} 
\definecolor{iccvblue}{rgb}{0.21,0.49,0.74}
\usepackage[pagebackref,breaklinks,colorlinks,allcolors=iccvblue]{hyperref}

\def\paperID{13010} 
\def\confName{ICCV}
\def\confYear{2025}

\title{Towards a 3D Transfer-based Black-box Attack via Critical Feature Guidance}

\author{Shuchao Pang\textsuperscript{1}\footnotemark[1]
\quad Zhenghan Chen\textsuperscript{2}\footnotemark[1]
\quad Shen Zhang\textsuperscript{1}
\quad Liming Lu\textsuperscript{1}\\
\quad Siyuan Liang\textsuperscript{3}
\quad Anan Du\textsuperscript{4}\footnotemark[2]
\quad Yongbin Zhou\textsuperscript{1}\\
\textsuperscript{1} Nanjing University of Science and Technology \quad 
\textsuperscript{2} STCA, Microsoft\\
\textsuperscript{3} Nanyang Technological University \quad 
\textsuperscript{4} Nanjing University of Industry Technology\\
\small \texttt{\{pangshuchao, zhangshen, luliming, zhouyongbin\}@njust.edu.cn,}\\
\small \texttt{zhenghan.chen@alumni.pku.edu.cn, siyuan.liang@ntu.edu.sg, anan.du@niit.edu.cn}
}

\begin{document}
\maketitle

\footnotetext[1]{These authors contributed equally.}
\footnotetext[2]{Corresponding author.}

\begin{abstract} \label{abstract}
  Deep neural networks for 3D point clouds have been demonstrated to be vulnerable to adversarial examples. Previous 3D adversarial attack methods often exploit certain information about the target models, such as model parameters or outputs, to generate adversarial point clouds. However, in realistic scenarios, it is challenging to obtain any information about the target models under conditions of absolute security. Therefore, we focus on transfer-based attacks, where generating adversarial point clouds does not require any information about the target models. 
  Based on our observation that the critical features used for point cloud classification are consistent across different DNN architectures, we propose CFG, a novel transfer-based black-box attack method that improves the transferability of adversarial point clouds via the proposed \textbf{C}ritical \textbf{F}eature \textbf{G}uidance. Specifically, our method regularizes the search of adversarial point clouds by computing the importance of the extracted features, prioritizing the corruption of critical features that are likely to be adopted by diverse architectures. Further, we explicitly constrain the maximum deviation extent of the generated adversarial point clouds in the loss function to ensure their imperceptibility. 
  Extensive experiments conducted on the ModelNet40 and ScanObjectNN benchmark datasets demonstrate that the proposed CFG outperforms the state-of-the-art attack methods by a large margin.
  The code is available at \url{https://github.com/AIASLab/CFG-ICCV2025}.
\end{abstract}

\section{Introduction} \label{Introduction}

Recent advancements \cite{lu2024uniads, xu2024enhancing, pang2025pridm, wang2024dp} in deep learning have significantly impacted 3D perception systems, especially in safety-critical areas such as autonomous driving and robotics \cite{xu2024memory, jiang2023symphonize, huang2023ptt}. However, the vulnerability of 3D perception models to adversarial attacks raises serious security concerns. While adversarial attacks on 2D images are well explored \cite{liu2024hqa,wu2024towards,liu2024explicitly,65, cheng2023topology,kong2024patch,liang2021generate,liang2020efficient,wei2018transferable,liang2022parallel,liang2022large}, adversarial robustness in 3D remains an emerging research area, with unique challenges stemming from the unordered, sparse, and structured nature of point cloud data.

\begin{figure}[htb]
    \centering
    \includegraphics[width=\linewidth]{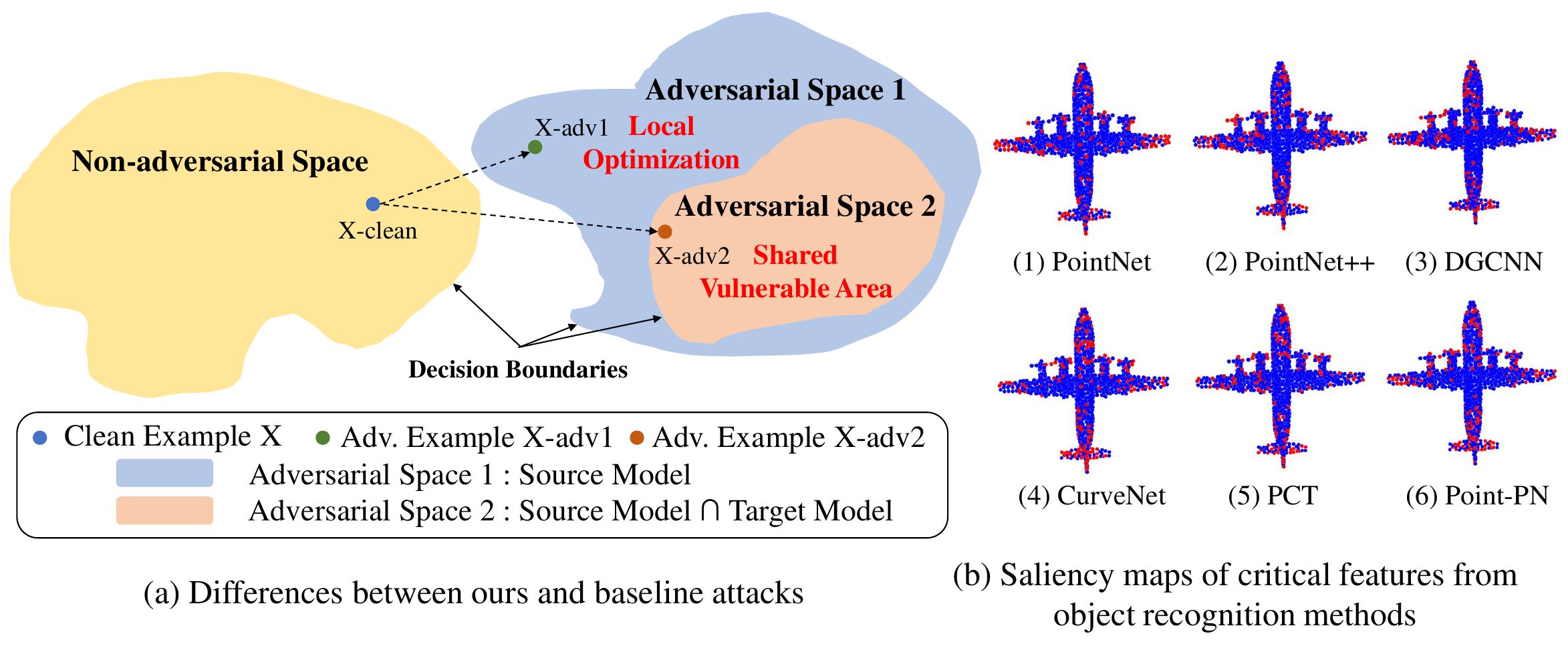}
    \caption{Illustration of our motivations. In (a), X-clean is the clean point clouds, X-adv1 and X-adv2 are the adversarial (\textit{abbr.}: adv.) point clouds generated by baseline attacks and our attack, where source\&target models can predict correctly in Non-adv. Space, the source model and the target model output wrong predictions in Adv. Space 1\&2 and only in Adv. Space 2, respectively. Overall, X-adv1 in Adv. Space 1 is a local optimization adv. sample because the target model still predicts correctly. In contrast, X-adv2 from our method does due to it being located in a shared vulnerable area (i.e., Adv. Space 2) from both source \& target models. In (b), we observe that existing DNN models focus on the same position features that are critical for classification (More samples are in supplementary).}
    \label{fig:1}
\end{figure}

Most existing adversarial attack studies on 3D point clouds focus on the white-box setting, where the attacker has full access to the model's parameters and gradients \cite{xiang2019generating,wen2020geometry}. However, in practical scenarios, such access is often unavailable, making black-box attacks significantly more relevant and challenging, where the attacker has no access to the target model’s parameters or gradients.  A few studies have attempted black-box attacks, but they either rely on query-based methods (which require numerous queries to approximate the model’s gradients) or suffer from low attack transferability \cite {60, tao20233dhacker}.
One promising approach for black-box attacks is transfer-based black-box attacks, where adversarial examples crafted on a source model are transferred to an unseen target model. Although transfer-based attacks have been widely studied in the 2D domain, their effectiveness in 3D remains highly limited due to the irregular and unordered nature of point cloud representations. Therefore, this work aims to develop more effective and generalizable transfer-based black-box attack techniques. 

\textcolor{black}{One intuitive idea is to generate the adversarial point clouds directly on the source model using white-box attack methods, however, the crafted adversarial point clouds often exhibit weak transferability due to overfitting to the source model. For example, as shown in \figurename~\ref {fig:1}(a), under white-box attacks, X-clean can easily enter the Adversarial Space 1 of the source model as X-adv1, but it cannot ensure enter the Adversarial Space 2 of the common space between source and target models as X-adv2 and thus cannot pose a threat to the target model.} So the key to the transfer-based attack is to reduce the overfitting of the adversarial point clouds to the source model. Consequently, very recent research efforts have been directed towards mitigating this overfitting issue by incorporating additional operations into the optimization process, e.g., AdvPC \cite{hamdi2020advpc} is to jointly attack the original input point clouds and the point clouds reconstructed by an autoencoder, so that it will rely less on the source model and better generalize to target models. PF-Attack \cite{he2023generating} introduces the concept of sub-perturbations, which regularizes the search for adversarial perturbations through the valid information of sub-perturbations. These attack methods generate adversarial point clouds by optimizing objectives associated with the network's softmax output. However, existing attack methods would indiscriminately destroy features that may contain some features unique to the source model, reducing the transferability of the adversarial point clouds. 


\textit{Observations:} For DNNs-based models in identifying point clouds, we observe that the key points that play a crucial role in recognition across different 3D deep models exhibit significant overlap and tend to focus on regions corresponding to the typical parts associated with the object category, like the nose, wings and tail of an airplane, as shown in \figurename~\ref{fig:1}(b).  We refer to their corresponding features as critical features and leverage these critical features to guide the direction of our transfer-based black-box attack on 3D adversarial point clouds.

Motivated by the above observations and \textcolor{black}{some work in 2D images \cite{ganeshan2019fda, wang2021feature,wu2020boosting}}, we propose a Critical Feature Guidance (CFG) attack, a transfer-based black-box attack approach, which significantly improves the transferability of the adversarial point cloud guided through an iterative search in the direction of shared vulnerable area between source and target models. Specifically, we first utilize the gradient information to evaluate the important distribution of features and assign high weights to critical features to achieve their precise destruction. 
Then, we construct a novel loss function that incorporates multiple objectives such as transferability, imperceptibility, and classification error, and finally use an iterative search technique to find the optimal adversarial point clouds in the adversarial space.

In summary, we highlight the following contributions of this work:

\begin{itemize}
    \item We observe the critical features employed are consistent across DNN architectures for point cloud recognition, which provides direction for finding shared vulnerable areas among source and target models and helps generate more aggressive 3D adversarial point clouds.

    \item Based on critical feature guidance, we propose a novel transfer-based black-box attack method, called CFG, which significantly improves the transferability of adversarial point clouds by prioritizing the destruction of critical features.

    \item Comprehensive experiments conducted on the benchmark datasets verify the high effectiveness of our CFG attack compared to the state-of-the-art methods and its resistance to potential defenses.

\end{itemize}

\section{Related Work} \label{Related Work}

\subsection{Point-Based DNNs for 3D Point Cloud Classification}
In recent years, deep point cloud learning \cite{67,68,69,70} has emerged with diverse applications in many fields, where DNNs-based models can be categorized as follows. 
\textit{PointNet and its Series}: PointNet \cite{qi2017pointnet} has been proposed as a pioneering and efficient framework, and then PointNet++ \cite{qi2017pointnet++} further improves local feature repsentation by dividing point clouds into subregions with two forms: single-scale (SSG) and multi-scale (MSG). And DGCNN \cite{wang2019dynamic} designs the EdgeConv module for better extracting local features on point clouds.  \textit{Convolution/Curve-based Methods}: PointConv \cite{wu2019pointconv} is proposed to efficiently extract local features using an irregular mesh convolution operation. CurveNet \cite{xiang2021walk} develops a long-range point cloud feature extraction operator by utilizing geometric information and a series of curves. \textit{Transformer-based Methods}: With the revolution of transformer in 2D visual tasks, PCT \cite{53} is proposed by incorporating a self-attention mechanism of transformer into point cloud recognition to extract contextual features effectively. \textit{Non-parametric-based Methods}: Further, Point-NN \cite{59} is proposed without using any learnable operators and on top of it, Point-PN \cite{59} is designed by inserting linear layers as a lightweight and efficient approach for 3D object recognition.

\subsection{Adversarial Attacks against Point-Based DNNs}

\begin{figure*}[ht]
    \centering
    \includegraphics[width=0.8\linewidth]{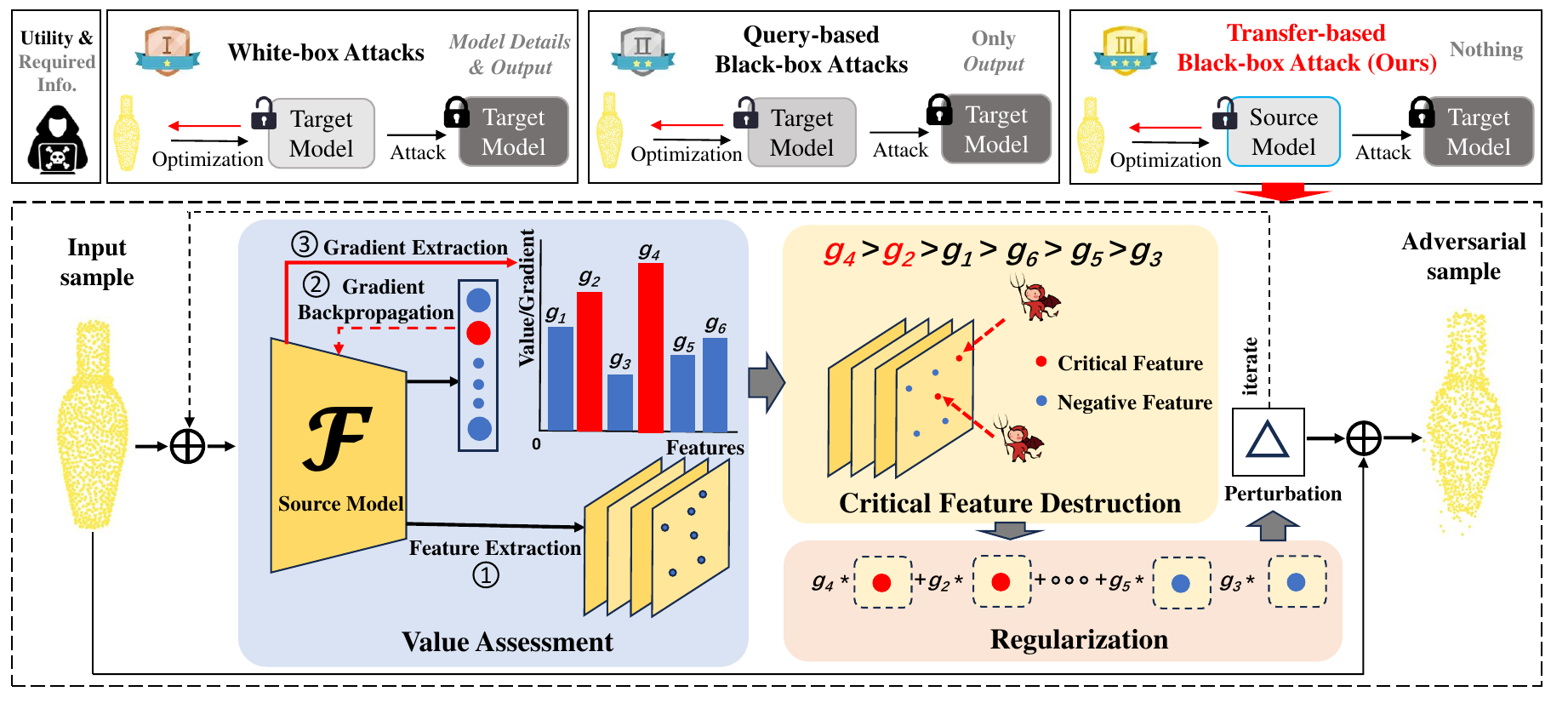}
    \caption{Overview of our proposed CFG method and its differences with baseline attacks.}
    \label{fig:2}
\end{figure*}

\textbf{3D White-box Attacks:} 3D-Adv \cite{xiang2019generating} is the first work to investigate the generation of adversarial point clouds by point perturbation and adding additional points. Liu et al. \cite{13} utilize the fast gradient sign method and utilize clipping to constrain and generate 3D adversarial point clouds.  KNN \cite{tsai2020robust} incorporates KNN distance into the loss function to generate reasonably shaped adversarial point clouds. GeoA3 \cite{wen2020geometry} gives a geometry-aware loss term to enhance the unpredictability of adversarial point clouds to humans. However, as shown in \figurename~\ref{fig:2}, all these white-box attack methods require knowledge of all target model parameters, and they are rarely used in real scenarios. 

\noindent \textbf{3D Black-box Attacks:} For \textit{Query-based Attacks}, Huang et al. \cite{60} propose the point-cloud sensitivity map to limit point perturbations on shaped surfaces. For \textit{Transfer-based Attacks}, 
AdvPC attack \cite{hamdi2020advpc} is proposed to jointly attack the original input point clouds and reconstruct the point clouds by an autoencoder. Liu et al. \cite{liu2022boosting} propose a point cloud frequency attack method, named AOF attack, to attack the more general features of the 3D point cloud, thereby improving the transferability of the 3D adversarial point clouds. PF-Attack \cite{he2023generating} is proposed to regularize the search for adversarial perturbations. However, these black-box attacks do not focus on the differences in internal features of models, which is critical to improve the transferability of the adversarial point clouds.

\subsection{Adversarial Defenses towards Point-Based DNNs}

\textbf{Statistical Outlier Removal:} This kind of method aims to preprocess the original point clouds before feeding into models to remove as many outliers as possible. SRS \cite{34} is a widely used strategy first to downsample the point clouds into a smaller set and then take the predicted class of such set as the result of the original point clouds. Further, DUP-Net \cite{16} utilizes Statistical Outlier Removal (i.e., SOR) operations to identify and remove outliers from the point clouds. IF-Defense \cite{22}is designed to optimize and restore input points by restricting the extent of point perturbations and surface deformation coordinates. 

\noindent \textbf{Robust Model Architecture:} Very few studies focus on designing robust model architecture to defend against various adversarial sample attacks. Li et al. \cite{39} introduce CCN, a novel point cloud architecture that can smooth and disrupt adversarial perturbations. 

\noindent \textbf{Data Augmentation:}  Data augmentation aims to increase the number of training samples for improving the model's defense capability. Liu et al. \cite{13} explore the use of adversarial point clouds in the 3D domain, aiming to enhance the adversarial robustness of models. PointCutMix \cite{21} involves pairing two training point clouds and exchanging a set number of points, creating an entirely new, mixed dataset. Li et al. \cite{39} propose a unique data augmentation strategy that adaptively balances the model's usability with its robustness for adversarial attacks.



\section{METHODOLOGY} \label{METHODOLOGY}

\subsection{Preliminaries} 

The aim of adversarial attacks for point cloud data is to mislead a 3D target model to misclassify an adversarial point cloud. 
Formally, for a classification task, 
let $x$ be an arbituary clean point cloud with true label $y$, and $\mathcal{G}_\theta$ denote the target model with parameters $\theta$, then $\mathcal{G}_\theta(x) = y$. The purpose of the adversarial attack by perturbing is to generate an imperceptible adversarial point cloud $x^{adv} = x + \boldsymbol{\Delta}$, where $\boldsymbol{\Delta}$ denotes perturbation, to make $\mathcal{G}_\theta(x^{adv})$ produce a wrong prediction, that is 

\begin{equation} \label{eq1}
\min \mathcal{D}(x^{adv}-x), \ \ s.t. \  \mathcal{G}_\theta(x^{adv}) \neq y,
\end{equation} 
where $\mathcal{D}$ is a distance metric.
For simplicity, the above problem can be reformulated as the following optimization problem:

\begin{equation} \label{eq1_reform}
\min _{x^{a d v}} - \mathcal{L}_{cls}\left(\mathcal{G}_\theta\left(x^{a d v}\right), y\right), \ \ s.t. \  \mathcal{D}(x^{adv}-x) \leq \epsilon,
\end{equation}  
where, $- \mathcal{L}_{cls}$ is a loss term used to facilitate the target model $\mathcal{G}$ to misclassify $x^{adv}$, 
and $\epsilon$ is a fairly small constant to improve the imperceptibility of the generated adversarial point clouds.

\subsection{Problem Formulation}

In this paper, we challenge the more realistic transfer-based black-box attack, where no information about the target model can be obtained. Formally, we generate adversarial point clouds on a knowable source model $\mathcal{F}_\phi$, exploiting their transferability to fool the target model $\mathcal{G}_\theta$. Derived from Eq. (\ref{eq1_reform}), the transfer-based attack can be formulated as:

\begin{equation}
\begin{aligned}
 \min _{x^{a d v}} - \mathcal{L}_{cls}\left(\mathcal{F}_\phi\left(x^{a d v}\right), y\right) +  \alpha  \cdot \mathcal{L}_{reg}\left(x^{adv},x\right),\\ \ \ s.t. \  \mathcal{D}(x^{adv}-x) \leq \epsilon,
 \end{aligned} 
 \label{eq_attack}
\end{equation} 
where $\mathcal{L}_{reg}$ is a regularization term used to guide the search for adversarial point clouds towards fooling the source and the target models, and $\alpha \in (0,1]$ is the penalty parameter.

\subsection{Critical Feature Guidance}
The key to the transfer-based attack is to enhance the transferability of adversarial point clouds from the source model to the target model by designing a suitable regularization term $\mathcal{L}_{reg}$.

Through extensive experimental research,  we find that the key points that play a crucial role in recognition across different 3D deep models are quite similar and exhibit significant overlap. We refer to their corresponding features as critical features. 
Specifically, we first obtain the saliency maps from various 3D deep models for the recognition task referred to \cite{zheng2019pointcloud}. The saliency maps assign scores to all points in a point cloud based on their importance in recognition. The higher the score, the more important the point and vice versa. For the sake of clearer observation, we select the top 200 points with the highest scores and mark them in red, while the remaining points are marked in blue. We observe that various 3D models tend to assign higher scores to points in similar regions, and these regions correspond to the typical parts associated with the object category, e.g., the head, tail and wings of an airplane in \figurename~\ref{fig:1}(b). This observation indicates that, although the underlying architectures of different 3D models vary, their reliance on geometric structure remains consistent. Therefore, we exploit this consistency to improve the transferability of the adversarial point clouds from the source model to different target models. 

We propose a critical feature guidance strategy for the transfer-based attack. As shown in \figurename~\ref{fig:2}, the strategy is mainly divided into two steps: 1) value assessment and 2) critical feature destruction. 

\textbf{Value assessment} aims to rank features' importance to identify critical and negative features. \textcolor{black}{Since features' importance is proportional to how much the feature influences the final decision}, we use gradient backpropagation to get the gradient information $G$ of the features, and then we use the gradient as the basis of the importance of the features to get the critical features we expect. 

\textbf{Critical feature destruction} aims to destroy the shared vulnerable areas between the different models. \textcolor{black}{We typically target the softmax layer for attacks; however, the adversarial samples generated at this layer often preserve the high-level semantic information of the original samples, which can reduce the success rate of the attack \cite{ganeshan2019fda}. Therefore, we opt to target a middle layer for our attack.} After obtaining the critical features in the middle layer, we increase the weights of critical features so as to guide the search of the adversarial point clouds in the direction of critical feature destruction. 

Formally, our regularization term is designed as:

\begin{equation} \label{eq2}
\mathcal{L}_{CFG}\left(x^{a d v}, \mathcal{F}_\phi\right)=G \odot A_k^{\mathcal{F}_\phi}\left(x^{a d v}\right),
\end{equation}
where $A_k^{\mathcal{F}_\phi}\left(\cdot\right)$ represents the feature map of the $k$-th layer of the $\mathcal{F}_\phi$, critical features are given high weights. Then, the objective of the transfer-based attack, Eq.~(\ref{eq_attack}), is derived to:

\begin{equation}
\begin{aligned}
 \min _{x^{a d v}} - \mathcal{L}_{cls}\left(\mathcal{F}_\phi\left(x^{a d v}\right), y\right) +  \alpha  \cdot \mathcal{L}_{CFG}\left(x^{a d v}, \mathcal{F}_\phi\right), 
 \\ \ \ s.t. \  \mathcal{D}(x^{adv}-x) \leq \epsilon.
 \end{aligned} 
 \label{eq_cfg}
\end{equation}

\paragraph{Analysis:} Our proposed regularization term, which is based on critical feature guidance, can guide the search direction of the adversarial point clouds towards a direction that is more beneficial to the destruction of the critical features. Since various models rely heavily on these critical features, this strategy helps to reduce the overfitting of the generated adversarial point clouds to the source model, and thus improves their transferability to other target models.

\subsection{Generating adversarial point clouds}
Following \cite{hamdi2020advpc,he2023generating}, we use the $l_\infty$ norm as the distance metric $\mathcal{D}$, because it is the most widely used in the community. However, it is not enough to guarantee the imperceptibility of the generated adversarial point clouds with a single $l_\infty$ norm. Therefore, we add an additional distance constraint $\mathcal{L}_{CD}$, the Chamfer distance,  to the loss function. We use the Chamfer distance because it is insensitive to the order of points and is suitable for unordered point cloud data. 

Overall, combining Eq.~(\ref{eq2}) and Eq.~(\ref{eq_cfg}), the objective of our transfer-based attack via critical feature guidance is derived as 

\begin{multline}
\begin{aligned}
 \min _{x^{a d v}} - \mathcal{L}_{cls}\left(\mathcal{F}_\phi\left(x^{a d v}\right), y\right) +  \alpha  \cdot  \mathcal{L}_{CFG}\left(x^{a d v}, \mathcal{F}_\phi\right) + \\ \beta \cdot \mathcal{L}_{CD}(x^{a d v}, x)  ], 
 \ \ s.t. \ \left\|x-x^{a d v}\right\|_\infty \leq \epsilon,
\end{aligned} 
\label{eq6}
\end{multline} 
where $\beta \in (0,10]$ is a hyper-parameter to balance the constraints.
Algorithm \ref{alg:algorithm} specifically shows the process of generating adversarial point clouds.

\begin{algorithm}[h]
\caption{Critical Feature Guidance Attack}
\label{alg:algorithm}
\textbf{Input}: Point cloud $x$, true label $y$, Unit normal vector of benign point cloud $n_x$ , number of iteration steps $T$, step size $\eta$, optimizer Adam, coefficient of $l_\infty$ norm $\epsilon$;


\begin{algorithmic}[1] 
\STATE Initialize Perturbation $\boldsymbol{\Delta}_0 \sim N(0,0.001)$;
\STATE $x_{0}^{a d v} \leftarrow x + \boldsymbol{\Delta}_0$;
\STATE $t \leftarrow 0$;
\WHILE{$t \leq T$}
    \STATE $\boldsymbol{\Delta}_{t+1} \leftarrow \boldsymbol{\Delta}_t-\eta \cdot \nabla \mathcal{L}\left(x, x^{a d v}, y, \mathcal{F}_\phi\right)$;
    \STATE $\boldsymbol{\Delta}_{t+1}^{\prime} \leftarrow\left\langle\boldsymbol{\Delta}_{t+1}, n_x\right\rangle \cdot n_x$;
    \IF{$\left|\boldsymbol{\Delta}_{t+1}^{\prime}\right|  \leq \epsilon $} 
    \STATE $x_{t+1}^{a d v} \leftarrow x+\boldsymbol{\Delta}_{t+1}^{\prime}$
    \ELSE
    \STATE $x_{t+1}^{a d v} \leftarrow x+\frac{\boldsymbol{\Delta}_{t+1}^{\prime}}{\left\|\boldsymbol{\Delta}_{t+1}^{\prime}\right\|_2} \cdot \epsilon$
    \ENDIF 
\STATE $t \leftarrow t + 1$;
\ENDWHILE
\STATE \textbf{Output}: $x_{t+1}^{a d v}$
\end{algorithmic}
\end{algorithm}



\section{Experiments}
\subsection{Experimental Settings}
\label{Experimental Settings}



\paragraph{Datasets}
We do experiments on two widely used point cloud benchmark datasets ModelNet40 \cite{wu20153d} and ScanObjectNN \cite{uy2019revisiting}.
\textbf{i) ModelNet40} is a synthetic CAD model dataset that contains 12,311 objects from 40 most common object categories. Among them, 9,843 objects are used for training and the other 2,468 for testing. We uniformly sample 1024 points from the surface of each object to form the input point cloud as \cite{4}. Follow \cite{xiang2019generating, liu2022imperceptible, he2023generating}, we evaluate the adversarial point cloud attacks on a subset of the testing set, containing 25 instances per category for 10 selected object categories. \textbf{ii) ScanObjectNN} is a real-world scanned dataset that contains 15,000 objects extracted from real-world scans, which are categorized into 15 classes. Due to the existence of background, noise, and occlusions, this benchmark poses significant challenges to existing methods.

\begin{figure}[ht]
\begin{minipage}[]{0.44 \textwidth}
\centering
\includegraphics[width=\textwidth]{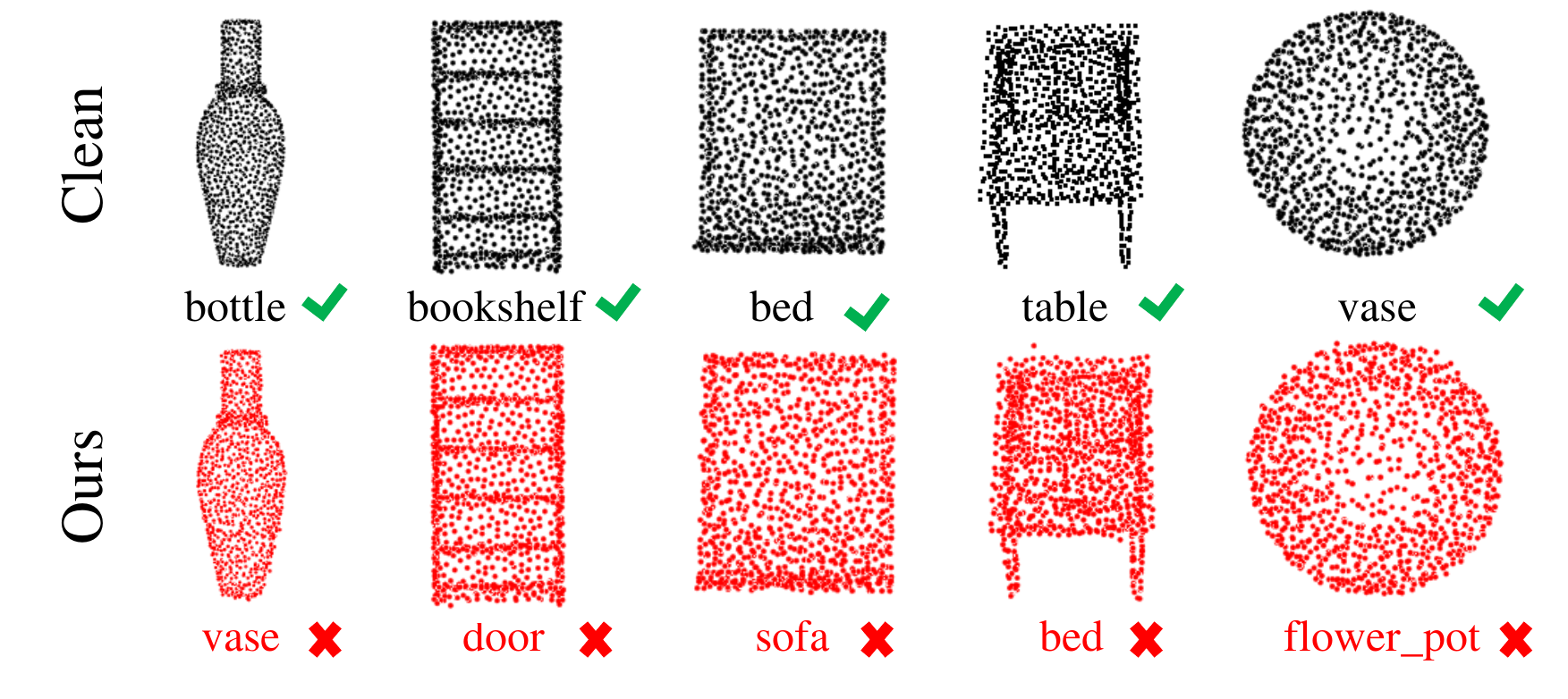}
\caption{Visualization of clean/adversarial samples generated by our method on ModelNet40.}
\label{fig:4}
\end{minipage}
\hfill
\begin{minipage}[]{0.44\textwidth}
\centering
\includegraphics[width=\textwidth]{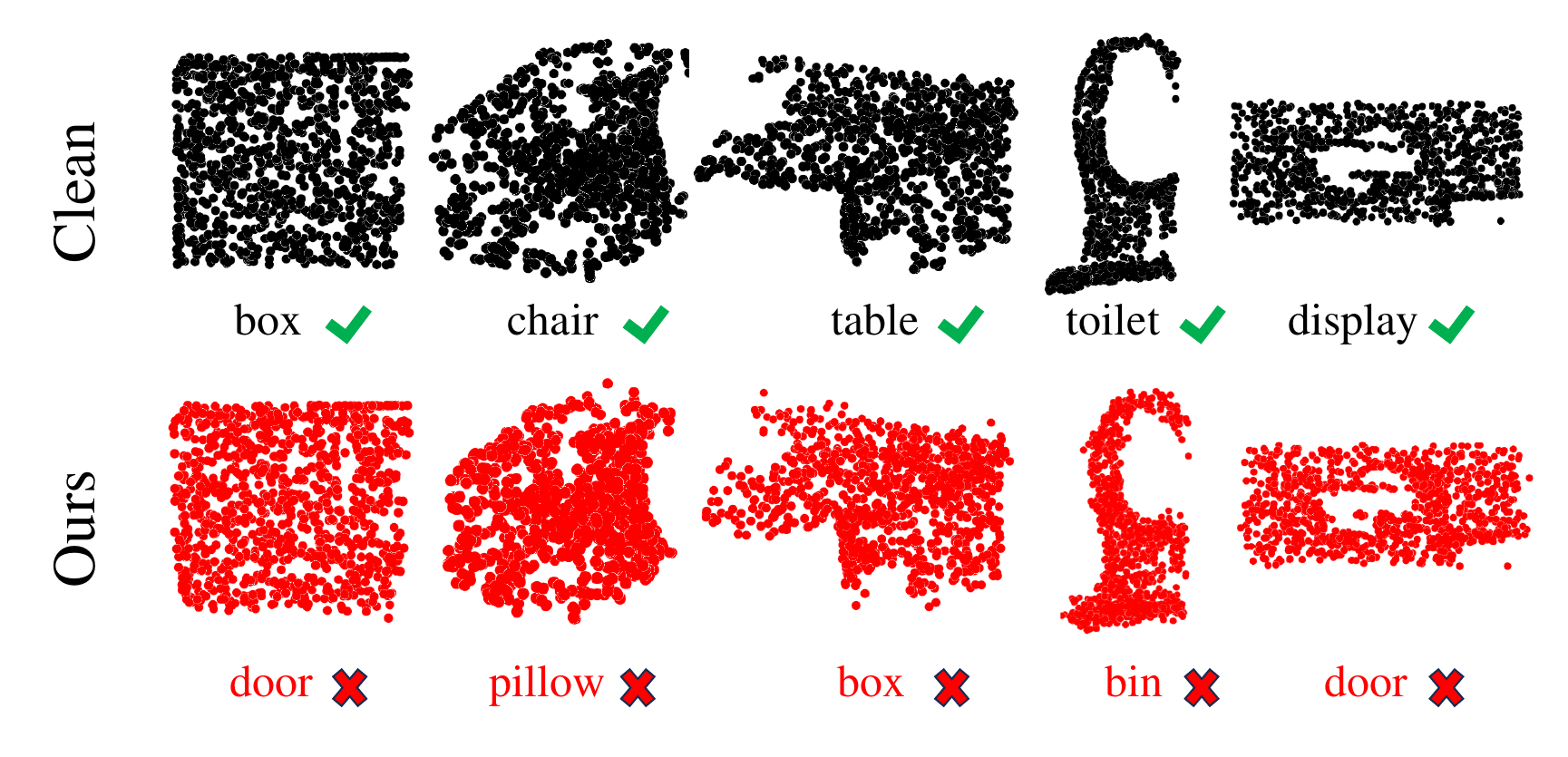}
\caption{Visualization of clean/adversarial samples generated by our method on ScanObjectNN.}
\label{fig:visual_scan}
\end{minipage}
\end{figure}

\paragraph{Source\&Target Models and Comparison Methods}

Following previous works \cite{hamdi2020advpc,he2023generating}, we select four classical point cloud classification models, i.e., PointNet \cite{4}, PointNet++ (MSG/SSG)\cite{qi2017pointnet++} and DGCNN \cite{wang2019dynamic}, and alternately use them as the source models and target models to evaluate the effectiveness of our attack method. Moreover, as advanced models typically increase the difficulty of attacks, we also evaluate the transferability of our proposed attack on the state-of-the-art target models, i.e., PointConv \cite{wu2019pointconv}, PointCNN \cite{li2018pointcnn}, CurveNet \cite{xiang2021walk}, PCT \cite{53}, PT \cite{zhao2021point} and Point-PN \cite{59}.
Further, to demonstrate the effectiveness of the proposed CFG, we compare with five widely-used attack methods: 3D-Adv \cite{xiang2019generating}, KNN \cite{tsai2020robust}, GeoA3 \cite{wen2020geometry}, AdvPC \cite{hamdi2020advpc} and PF-Attack \cite{he2023generating}.

\paragraph{Implementation Details}

The hyper-parameters of the attack are set to:
We set the step size $\eta = 0.01$, the number of iterations for the attack optimization for all the networks $T = 200$ and use the Adam optimizer. Following previous works \cite{hamdi2020advpc,he2023generating}, we set $\epsilon \in \left\{0.18, 0.45\right\}$, balance hyper-parameters $\beta = 10$. We choose the penalty parameter $\alpha = 0.8$ based on the ablation study. All attacks initialize the perturbation once and limit the perturbation size using $\epsilon_\infty$. 8 RTX 3090 GPU cards are used for calculation.

\begin{table*}[tb]
\centering
\caption{Attack transferability across classical models on ModelNet40. Measure performance in terms of attack success rate (\%). The results of 3D-Adv, KNN, AdvPC and PF-Attack are reported in \cite{he2023generating}. The number in bold indicates the best.}
\label{tab:res_modelnet_classical_multi}
\resizebox{\textwidth}{!}
{%
\begin{tabular}{cc*{4}{>{\centering\arraybackslash}m{0.13\linewidth}} | *{4}{>{\centering\arraybackslash}m{0.13\linewidth}}}
\hline
\multirow{2}{*}{\begin{tabular}[c]{@{}c@{}} Source\\ Model\end{tabular}} &
  \multirow{2}{*}{\begin{tabular}[c]{@{}c@{}}Attack\\ Method\end{tabular}} &
  \multicolumn{4}{c|}{$\epsilon$ = 0.18} &
  \multicolumn{4}{c}{$\epsilon$ = 0.45} \\ \cline{3-10} 
 &
   &
  PointNet &
  \begin{tabular}[c]{@{}c@{}}PointNet++\\ (MSG)\end{tabular} &
  \begin{tabular}[c]{@{}c@{}}PointNet++\\ (SSG)\end{tabular} &
  DGCNN &
  PointNet &
  \begin{tabular}[c]{@{}c@{}}PointNet++\\ (MSG)\end{tabular} &
  \begin{tabular}[c]{@{}c@{}}PointNet++\\ (SSG)\end{tabular} &
  DGCNN \\ \hline
\multirow{6}{*}{PointNet} &
  3D-Adv \cite{xiang2019generating} &
  $100$ &
  8.4 &
  10.4 &
  6.8 &
  $100$ &
  8.8 &
  9.6 &
  8.0 \\
 &
  KNN \cite{tsai2020robust} &
  $100$ &
  9.6 &
  10.8 &
  6.0 &
  $100$ &
  9.6 &
  8.4 &
  6.4 \\
 &
  GeoA3 \cite{wen2020geometry} &
  $100$ &
  20.0 &
  19.6 &
  7.2 &
  $100$ &
  23.6 &
  20.8 &
  7.2 \\
 &
  AdvPC \cite{hamdi2020advpc} &
  $98.8$ &
  20.4 &
  27.6 &
  22.4 &
  $98.8$ &
  18.0 &
  26.8 &
  20.4 \\
 &
  PF-Attack \cite{he2023generating}&
  $100$ &
  49.6 &
  61.5 &
  24.8 &
  $100$ &
  59.3 &
  64.8 &
  28.8 \\
 &
  Ours &
  $100$ &
  \textbf{72.6} &
  \textbf{80.6} &
  \textbf{39.3} &
  $100$ &
  \textbf{84.7} &
  \textbf{88.0} &
  \textbf{49.3} \\ \hline
\multirow{6}{*}{\begin{tabular}[c]{@{}c@{}}PointNet++\\ (MSG)\end{tabular}} &
  3D-Adv \cite{xiang2019generating}&
  6.8 &
  100 &
  28.4 &
  11.2 &
  7.2 &
  $100$ &
  29.2 &
  11.2 \\
 &
  KNN \cite{tsai2020robust}&
  6.4 &
  $100$ &
  22.0 &
  8.8 &
  6.4 &
  $100$ &
  23.2 &
  7.6 \\
 &
  GeoA3 \cite{wen2020geometry}&
  4.4 &
  $100$ &
  14.4 &
  6.4 &
  4.4 &
  $100$ &
  13.6 &
  6.0 \\
 &
  AdvPC \cite{hamdi2020advpc}&
  13.2 &
  $97.2$ &
  54.8 &
  \textbf{39.6} &
  18.4 &
  $98$ &
  58.0 &
  \textbf{39.2} \\
 &
  PF-Attack \cite{he2023generating}&
  17.2 &
  $100$ &
  67.0 &
  24.3 &
  19.2 &
  $100$ &
  75.0 &
  27.5 \\
 &
  Ours &
  \textbf{17.6} &
  $100$ &
  \textbf{72.9} &
  25.3 &
  \textbf{36.6} &
  $100$ &
  \textbf{83.5} &
  34.7 \\ \hline
\multirow{6}{*}{\begin{tabular}[c]{@{}c@{}}PointNet++\\ (SSG)\end{tabular}} &
  3D-Adv \cite{xiang2019generating}&
  7.6 &
  9.6 &
  $100$ &
  6.0 &
  7.2 &
  10.4 &
  $100$ &
  7.2 \\
 &
  KNN \cite{tsai2020robust}&
  6.4 &
  9.2 &
  $100$ &
  6.4 &
  6.8 &
  7.6 &
  $100$ &
  6.0 \\
 &
  GeoA3 \cite{wen2020geometry}&
  5.2 &
  10.4 &
  $100$ &
  2.2 &
  4.8 &
  9.2 &
  $100$ &
  4.0 \\
 &
  AdvPC \cite{hamdi2020advpc}&
  12.0 &
  27.2 &
  $99.2$ &
  22.8 &
  14.0 &
  30.8 &
  $99.2$ &
  27.6 \\
 &
  PF-Attack \cite{he2023generating}&
  13.9 &
  47.4 &
  $100$ &
  19.6 &
  15.7 &
  56.9 &
  $100$ &
  23.0 \\
 &
  Ours &
  \textbf{18.1} &
  \textbf{84.7} &
  $100$ &
  \textbf{38.3} &
  \textbf{37.7} &
  \textbf{86.4} &
  $100$ &
  \textbf{45.9} \\ \hline
\multirow{6}{*}{DGCNN} &
  3D-Adv \cite{xiang2019generating}&
  9.2 &
  11.2 &
  31.2 &
  $100$ &
  9.6 &
  12.8 &
  30.4 &
  $100$ \\
 &
  KNN \cite{tsai2020robust}&
  7.2 &
  9.6 &
  14.0 &
  $99.6$ &
  6.8 &
  10.0 &
  11.2 &
  $99.6$ \\
 &
  GeoA3 \cite{wen2020geometry}&
  4.4 &
  27.2 &
  27.6 &
  $100$ &
  4.4 &
  26.8 &
  25.6 &
  $100$ \\
 &
  AdvPC \cite{hamdi2020advpc}&
  19.6 &
  46.0 &
  64.4 &
  $94.8$ &
  32.8 &
  48.8 &
  64.4 &
  $97.2$ \\
 &
  PF-Attack \cite{he2023generating}&
  21.3 &
  60.9 &
  74.8 &
  $100$ &
  26.1 &
  79.7 &
  85.8 &
  $100$ \\
 &
  Ours &
  \textbf{29.3} &
  \textbf{70.1} &
  \textbf{88.9} &
  $100$ &
  \textbf{39.0} &
  \textbf{86.0} &
  \textbf{91.8} &
  $100$ \\ \hline
\end{tabular}%
}
\end{table*}

\paragraph{Evaluation Metrics}\label{Evaluation Metrics}
For each point cloud of the testing set, the attack methods will generate a corresponding adversarial point cloud. We use two metrics for evaluation:
\textbf{i) Attack success rate (ASR):} we use the attack success rate on the target model to evaluate the attack transferability of the adversarial point clouds. ASR is calculated as the proportion of samples that an attack method successfully causes a target model to misclassify. 
\textbf{ii) Transferability score:} to evaluate the overall transferability, following \cite{hamdi2020advpc,he2023generating}, we calculate the average ASRs of the adversarial point clouds generated by attack methods on target models and use this as the transferability score.

\subsection{Evaluation on ModelNet40 Dataset}
\paragraph{Attack Transferability across Classical Models}

To comprehensively evaluate the performance of our attack method, we conduct a series of comparative experiments based on four classical models. Specifically, every model is alternatively used as the source model, with the remaining models serving as the target models. 
Table \ref{tab:res_modelnet_classical_multi} reports the attack success rates (ASRs) with different source models of our method compared to five popular attack methods. We observe that our attack method achieves the highest ASRs in most cases, which demonstrates that our attack method performs better in transferability.
In particular, with the PointNet++ (MSG) and PointNet++ (SSG) as the target models, our method consistently achieves an attack success rate of more than 70\%, achieving a substantial lead over the other attack methods. 


\textit{Transferability \textit{v.s.} Imperceptibility.} The parameter $\epsilon$ limits the maximum deviation extent of the generated adversarial point clouds from the corresponding clean samples and, therefore, represents the imperceptibility of the attack methods. In general, creating highly transferable adversarial samples will make the adversarial samples more detectable and less imperceptible. We observe that all of the attack 
\begin{table}
\centering
\caption{Transferability score on ModelNet40.}
\label{tab:res_modelnet_score}
\resizebox{\linewidth}{!}
{%
\begin{tabular}{c *{6}{>{\centering\arraybackslash}m{0.2\linewidth}}}
\hline
\multirow{2}{*}{$\epsilon$} & \multicolumn{6}{c}{Transferability Score (\%)}           \\ \cline{2-7} 
                   & 3D-Adv \cite{xiang2019generating} & KNN \cite{tsai2020robust}& GeoA3 \cite{wen2020geometry}& AdvPC \cite{hamdi2020advpc}& PF-Attack \cite{he2023generating}& Ours          \\ \hline
0.18               & 12.2   & 9.7 & 12.4  & 30.8  & 40.2      & \textbf{53.1} \\
0.45               & 12.6   & 9.2 & 12.5  & 33.3  & 46.8      & \textbf{63.6} \\ \hline

\end{tabular}%
}
\end{table}
success rates increase when $\epsilon$ increases from $0.18$ to $0.45$, which supports this viewpoint. Therefore, we can make a trade-off between the transferability and imperceptibility of our attack method according to the practical requirements.

\textit{Overall transferability.}
We compare the overall transferability in Table \ref{tab:res_modelnet_score}. We observe that our method performs the best, leading by $12.9\%$ for $\epsilon = 0.18$ and $16.8\%$ for $\epsilon = 0.45$.

\paragraph{Attack Transferability on State-of-the-art Target Models}
We assess the effectiveness of our attack method against six state-of-the-art models compared to baseline attacks. As shown in Table \ref{tab:res_modelnet_sota}, our proposed transfer-based black-box attack method achieves a significant improvement relative to the baseline attack methods against the state-of-the-art models. Further, we find that the success rate of the attack on the PointConv model is exceptionally high, which might result from the similarity of the adversarial space between PointNet and PointConv.

\begin{table*}[tb]
\centering
\caption{Attack transferability on state-of-the-art models on ModelNet40 dataset. Measure performance in terms of attack success rate (\%). Adversarial point clouds are generated on PointNet.}
\label{tab:res_modelnet_sota}
\resizebox{\textwidth}{!}
{%
\begin{tabular}{c *{6}{>{\centering\arraybackslash}m{0.1\linewidth}} | *{6}{>{\centering\arraybackslash}m{0.1\linewidth}}}
\hline
\multirow{2}{*}{\begin{tabular}[c]{@{}c@{}}Attack\\ Method\end{tabular}} &
  \multicolumn{6}{c|}{$\epsilon$ = 0.18} &
  \multicolumn{6}{c}{$\epsilon$ = 0.45} \\ \cline{2-13} 
          & PointConv & PointCNN & CurveNet & PCT  & PT   & Point-PN & PointConv & PointCNN & CurveNet & PCT  & PT   & Point-PN \\ \hline
3D-Adv \cite{xiang2019generating}   & 0.8       & 1.0      & 1.8      & 0  & 0.2    & 1.2        & 0.8       & 0.1      & 1.8      & 0  & 1.0    & 0.7        \\
GeoA3 \cite{wen2020geometry}    & 10.3      & 2.1      & 5.5     & 3.3  & 18.3    & 5.5        & 14.9      & 5.5     & 6.7     & 7.5 & 19.6    & 10.2        \\
PF-Attack \cite{he2023generating}& 50.3      & 29.5     & 22.7     & 23.4 & 49.9 & 26.1     & 53.2      & 33.4     & 24.8     & 28.7 & 52.3 & 40.8     \\
Ours &
  \textbf{68.1} &
  \textbf{32.7} &
  \textbf{29.2} &
  \textbf{30.4} &
  \textbf{65.6} &
  \textbf{42.5} &
  \textbf{81.3} &
  \textbf{43.4} &
  \textbf{47.5} &
  \textbf{48.6} &
  \textbf{84.7} &
  \textbf{61.5} \\ \hline
\end{tabular}%
}

\end{table*}

\begin{table*}[tb]
\centering
\caption{Resistance to potential defenses  on ModelNet40 dataset. Measure performance in terms of attack success rate (\%). The source model is PointNet. The number in bold indicates the best.}
\label{tab:res_modelnet_defense}
\resizebox{\textwidth}{!}{%
\begin{tabular}{c*{4}{>{\centering\arraybackslash}m{0.15\linewidth}} | *{4}{>{\centering\arraybackslash}m{0.15\linewidth}}}
\hline
Attack    & \multicolumn{4}{c|}{$\epsilon$ = 0.18} & \multicolumn{4}{c}{$\epsilon$ = 0.45} \\ \cline{2-9} 
Method    & SRS \cite{34}  & SOR \cite{16}  & CCN \cite{39}  & AT \cite{13}   & SRS \cite{34}  & SOR \cite{16}  & CCN \cite{39}  & AT \cite{13}  \\ \hline
3D-Adv \cite{xiang2019generating}   & 26.4  & 17.2  & 0   & 5.0   & 1.0  & 17.2  & 0   & 1.0  \\
GeoA3 \cite{wen2020geometry}    & 58.3  & 43.8  & 11.25  & 3.8   & 53.8  & 40.8  & 11.7  & 3.8  \\
PF-Attack \cite{he2023generating}& 55.7  & 42.4  & 30.0  & 10.4  & 61.1  & 33.8  & 36.5  & 14.4 \\
Ours & \textbf{79.2} & \textbf{62.9} & \textbf{50.6} & \textbf{22.6} & \textbf{87.9} & \textbf{73.0} & \textbf{66.8} & \textbf{30.8} \\ \hline
\end{tabular}%
}
\end{table*}

\paragraph{Attacks Transferability against Potential Defenses}

To explore the ability of our method to resist defenses, we investigate the performance of attack methods against target models that used defenses. 
Here we select three common defenses in the field of point cloud security, respectively, i.e., Statistical Outliers Removal (e.g., SRS \cite{34} and SOR \cite{16}), Robust Model Architecture (e.g., CCN \cite{39}),
\begin{table}
\centering
\caption{Time comparison. The source model is PointNet. Running Time is the average time (in minutes) taken to generate one adversarial point cloud.}
\label{tab:ablation_time}
\resizebox{\linewidth}{!}
{%
\begin{tabular}{c|*{4}{>{\centering\arraybackslash}m{0.19\linewidth}}}
\hline
Attack Method & 3D-Adv \cite{xiang2019generating}& GeoA3 \cite{wen2020geometry}& PF-Attack \cite{he2023generating}& Ours \\ \hline
Running Time  & 0.04   & 0.05  & 0.10      & 0.07 \\ \hline
\end{tabular}%
}
\end{table}
 and Adversarial Training (e.g., AT \cite{13, liu2023exploring, lu2025adversarial, zhang2024lanevil, liang2023exploring}). As shown in Table \ref{tab:res_modelnet_defense}, our attack method still maintains an obvious attack success rate under these defenses and achieves an average improvement of 20 percentage points ahead of the baseline attacks. Specifically, for the AT defense method, although better than baseline methods, our attack success rate is not high due to the fact that AT improves its generalization ability by adding an adversarial point cloud to the target model during the training phase, which leads to an increase in the difficulty of searching for the adversarial point clouds.
\paragraph{Efficiency Analysis} \label{Efficiency Analysis}
In terms of time overhead (see Table \ref{tab:ablation_time}), although we have a slight gap with white-box attacks, the generation time of our adversarial point cloud is still less than that of the black-box attack, PF-Attack, guaranteeing a leading success rate of the transfer-based attack.

\begin{table}[tb]
\centering
\caption{Transfer-based attack on ScanObjectNN dataset. Measure performance in terms of attack success rate (\%). The source model is PointNet. The number in bold indicates the best.}
\label{subtab:res_scan_classical_abc}
\fontsize{10}{12}\selectfont
\begin{subtable}{\linewidth}
\centering
\caption{On Classical Target Models}
\label{subtab:res_scan_classical_PointNet}
\resizebox{\textwidth}{!}{%
\begin{tabular}{*{1}{>{\centering\arraybackslash}m{0.15\linewidth}} *{4}{>{\centering\arraybackslash}m{0.15\linewidth}} | *{4}{>{\centering\arraybackslash}m{0.15\linewidth}}}
\hline
 
  \multirow{2}{*}{\begin{tabular}[c]{@{}c@{}}Attack\\ Method\end{tabular}} &
  \multicolumn{4}{c|}{$\epsilon$ = 0.18} &
  \multicolumn{4}{c}{$\epsilon$ = 0.45} \\ \cline{2-9} 
   &
  PointNet &
  \begin{tabular}[c]{@{}c@{}}PointNet++\\ (MSG)\end{tabular} &
  \begin{tabular}[c]{@{}c@{}}PointNet++\\ (SSG)\end{tabular} &
  DGCNN &
  PointNet &
  \begin{tabular}[c]{@{}c@{}}PointNet++\\ (MSG)\end{tabular} &
  \begin{tabular}[c]{@{}c@{}}PointNet++\\ (SSG)\end{tabular} &
  DGCNN \\ \hline
  
  3D-Adv \cite{xiang2019generating}&
  100 &
  7.5 &
  7.5 &
  9.0 &
  100 &
  7.3 &
  7.5 &
  9.0 \\
 
  GeoA3 \cite{wen2020geometry}&
  100 &
  18.3 &
  20.1 &
  18.2 &
  100 &
  17.8 &
  19.5 &
  17.8 \\
 
  PF-Attack \cite{he2023generating}&
  100 &
  50.4 &
  54.6 &
  40.3 &
  100 &
  51.4 &
  54.2 &
  40.8 \\
 
  Ours &
  100 &
  \textbf{66.5} &
  \textbf{68.0} &
  \textbf{58.3} &
  100 &
  \textbf{69.1} &
  \textbf{70.3} &
  \textbf{58.4} \\ \hline
\end{tabular}%
}   
\end{subtable}
\begin{subtable}{\linewidth}
\centering
\caption{On State-of-the-art Target Models}
\label{subtab:res_scan_sota_PointNet}
\resizebox{\textwidth}{!}
{%
\begin{tabular}{*{1}{>{\centering\arraybackslash}m{0.15\linewidth}} *{6}{>{\centering\arraybackslash}m{0.1\linewidth}} | *{6}{>{\centering\arraybackslash}m{0.1\linewidth}}}
\hline
\multirow{2}{*}{\begin{tabular}[c]{@{}c@{}}Attack\\ Method\end{tabular}} &
  \multicolumn{6}{c|}{$\epsilon$ = 0.18} &
  \multicolumn{6}{c}{$\epsilon$ = 0.45} \\ \cline{2-13} 
          & Point-Conv & Point-CNN & Curve-Net & PCT  & PT & Point-PN & Point-Conv & Point-CNN & Curve-Net & PCT  & PT & Point-PN \\ \hline
3D-Adv \cite{xiang2019generating}   & 9.3     & 10.6      & 10.0     & 8.6  & 12.8  & 27.3        & 9.2       & 2.1      & 10.0     & 9.0  & 13.2  & 27.2        \\
GeoA3 \cite{wen2020geometry}    & 21.7      & 20.6     & 18.8      & 17.3  & 27.3  & 40.4        & 21.4      & 20.8     & 18.3      & 16.7  & 27.0  & 41.9        \\
PF-Attack \cite{he2023generating}& 50.1      & 45.0     & 47.1     & 45.6 & 51.9  & 68.8     & 47.9      & 45.5     & 46.4     & 47.1 & 51.8  & 69.5     \\
Ours &
  \textbf{61.7} &
  \textbf{50.9} &
  \textbf{64.5} &
  \textbf{63.6} &
  \textbf{67.0} &
  \textbf{80.0} &
  \textbf{63.8} &
  \textbf{53.7} &
  \textbf{66.5} &
  \textbf{66.8} &
  \textbf{68.9} &
  \textbf{81.8} \\ \hline
\end{tabular}%
}
\end{subtable}
\begin{subtable}{\linewidth}
\centering
\caption{Against Potential Defenses}
\label{subtab:res_scan_defense}
\resizebox{\textwidth}{!}{%
\begin{tabular}{*{1}{>{\centering\arraybackslash}m{0.15\linewidth}} *{4}{>{\centering\arraybackslash}m{0.15\linewidth}} | *{4}{>{\centering\arraybackslash}m{0.15\linewidth}}}
\hline
Attack    & \multicolumn{4}{c|}{$\epsilon$ = 0.18} & \multicolumn{4}{c}{$\epsilon$ = 0.45} \\ \cline{2-9} 
Method    & SRS \cite{34}  & SOR \cite{16}  & CCN \cite{39}  & AT \cite{13}   & SRS \cite{34}  & SOR \cite{16}  & CCN \cite{39}  & AT \cite{13}  \\ \hline
3D-Adv \cite{xiang2019generating}   & 67.4  & 67.8  & 8.3   & 23.5   & 48.7  & 48.7  & 8.3   & 23.5  \\
GeoA3 \cite{wen2020geometry}    & 77.6  & 59.4  & 23.4  & 27.5   & 77.2  & 59.0  & 22.8  & 26.9  \\
PF-Attack \cite{he2023generating}& 60.1  & 50.4  & 52.3  & 41.6  & 56.8  & 50.4  & 53.4  & 41.0 \\
Ours & \textbf{96.6} & \textbf{82.2} & \textbf{63.4} & \textbf{56.0} & \textbf{95.7} & \textbf{81.1} & \textbf{64.8} & \textbf{60.7} \\ \hline
\end{tabular}%
}    
\end{subtable}
\end{table}

\subsection{Evaluation on ScanObjectNN Dataset}
To verify the generalization of our attack method, we run similar experiments on the real dataset ScanObjectNN \cite{uy2019revisiting}, as shown in Table \ref{subtab:res_scan_classical_abc}. Here, we selected 3D-adv, GeoA3, and PF-Attack as baseline attack methods, and our method still outperforms all baseline attack methods by a large margin in improving the transferability of the adversarial point clouds. 
Notably, we find that all attack methods on the ScanObjectNN real dataset improve the transferability of the adversarial point clouds compared to ModelNet40.  This is not because the attack methods play a role but rather because the target model performs poorly with the challenging real dataset. More experimental results and analysis are shown in supplementary material.

\subsection{Ablation Studies}

\begin{table*}[tb]
\centering
\caption{Ablation studies of loss terms on ModelNet40 dataset. Measure performance in terms of attack success rate (\%). The attack success rate is reported by attacking other models using the deleted loss terms. The source model is PointNet. The number in bold indicates the best.}
\label{tab:ablation_loss}
\resizebox{\textwidth}{!}{%
\begin{tabular}{*{3}{>{\centering\arraybackslash}m{0.06\linewidth}}|*{4}{>{\centering\arraybackslash}m{0.13\linewidth}}|*{4}{>{\centering\arraybackslash}m{0.13\linewidth}}}
\hline
\multicolumn{3}{c|}{Loss Term} & \multicolumn{4}{c|}{$\epsilon$ = 0.18}                       & \multicolumn{4}{c}{$\epsilon$ = 0.45}                        \\ \hline
$\mathcal{L}_{cls}$ &
  $\mathcal{L}_{CD}$ &
  $\mathcal{L}_{CFG}$ &
  PointNet &
  \begin{tabular}[c]{@{}c@{}}PointNet++\\ (MSG)\end{tabular} &
  \begin{tabular}[c]{@{}c@{}}PointNet++\\ (SSG)\end{tabular} &
  DGCNN &
  PointNet &
  \begin{tabular}[c]{@{}c@{}}PointNet++\\ (MSG)\end{tabular} &
  \begin{tabular}[c]{@{}c@{}}PointNet++\\ (SSG)\end{tabular} &
  DGCNN \\ \hline
$\checkmark$        &          &          & 100 & 13.2          & 14.9          & 0.2           & 100 & 11.6          & 16.3          & 1.0           \\
$\checkmark$        & $\checkmark$        &          & 100 & 11.6          & 15.8          & 2.4           & 100 & 12.1          & 16.4          & 2.4           \\
$\checkmark$        &          & $\checkmark$        & 100 & \textbf{72.8} & \textbf{81.6} & 37.5          & 100 & 82.9          & 85.7          & \textbf{53.3} \\
$\checkmark$        & $\checkmark$        & $\checkmark$       & 100 & 72.6          & 80.6          & \textbf{39.3} & 100 & \textbf{84.7} & \textbf{88.0} & 49.3          \\ \hline
\end{tabular}%
}
\end{table*}


\begin{table}[tb]
\centering
\caption{\textcolor{black}{Comparison of imperceptibility on ModelNet40 dataset. The loss function is Eq. \ref{eq6} (with $\mathcal{L}_{CD}$ and without $\mathcal{L}_{CD}$). }}
\label{ablation_loss_cd}
\resizebox{\linewidth}{!}
{%
\begin{tabular}{c|ccc}
\hline
Loss       & Chamfer distance & Hausdorff distance & $\mathcal{L}_{2}$ distance \\ \hline
w. $\mathcal{L}_{CD}$    & 0.32             & 0.36               & 0.22          \\
w.o. $\mathcal{L}_{CD}$ & 0.40             & 0.40               & 0.25          \\ \hline
\end{tabular}%
}
\end{table}

\paragraph{Effectiveness of Different Loss Terms}

In our Eq.~(\ref{eq6}), based on $\mathcal{L}_{cls}$, two additional loss terms are analyzed in our ablation experiments to verify the validity of these parts of the proposed objective function, i.e., the imperceptible loss $\mathcal{L}_{CD}$ and the transferable loss $\mathcal{L}_{CFG}$. As shown in Table \ref{tab:ablation_loss}, when only the loss $\mathcal{L}_{CD}$ is considered, the transferability of the adversarial point clouds does not change significantly, and when only the loss $\mathcal{L}_{CFG}$ is considered, the transferability of the adversarial point clouds gets a significant performance improvement. Overall, the contribution of the loss $\mathcal{L}_{CFG}$ to the transferability is decisive, while the introduction of the loss $\mathcal{L}_{CD}$ can improve the imperceptibility (see Table \ref{ablation_loss_cd}) of the adversarial point clouds without affecting the transferability performance, and the combination of the two loss terms will result in the optimal adversarial point clouds.

\paragraph{Effect of Regularization Coefficient $\alpha$ on Attack Success Rates}

\begin{figure}[htb]
    \begin{minipage}[]{0.45\textwidth}
    \centering
    \includegraphics[width=\textwidth]{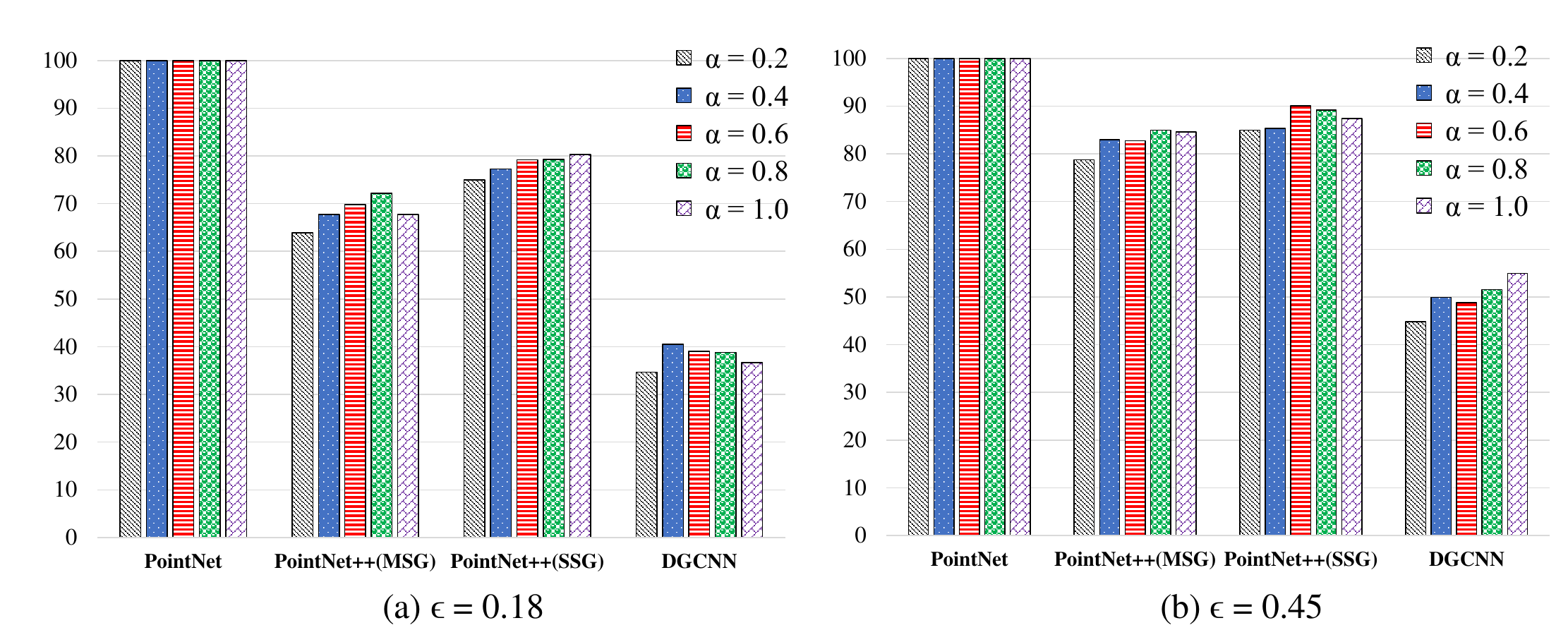}
    \caption{Ablation studies in $\alpha$. Measure performance in terms of attack success rate (\%). The source model is PointNet.}
    \label{fig:5}
    \end{minipage}
    \hfill
    \begin{minipage}[]{0.45\textwidth}
        \centering
        \includegraphics[width=\textwidth]{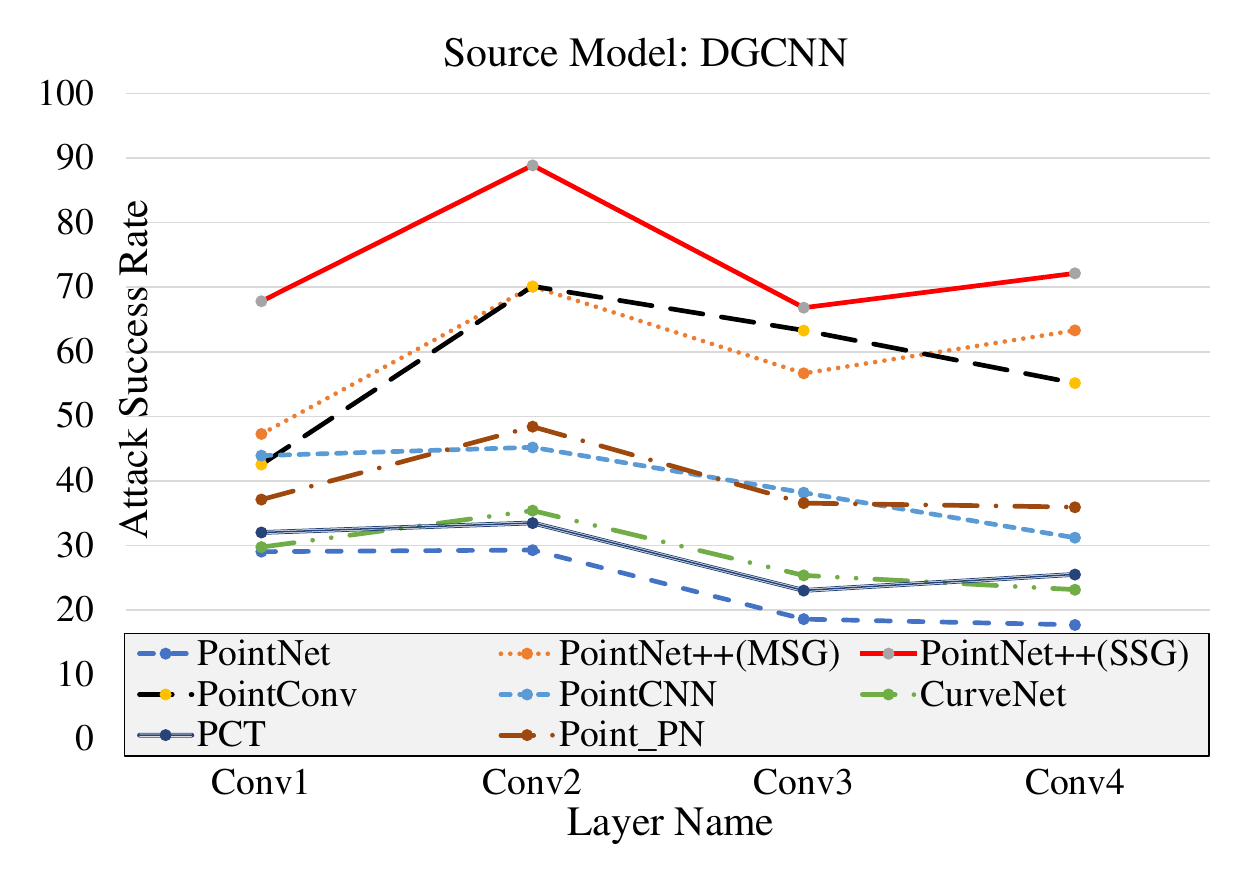}
        \caption{Effect of layer choice on attack success rate (\%). The source model is DGCNN. 
        }
        \label{fig:6}        
    \end{minipage}
\end{figure}

In this study, we investigate the effect of the regularization term $\alpha$ on the attack success rate, and we use PointNet as the source model to generate the adversarial point clouds. As shown in \figurename~\ref{fig:5}, different values of $\alpha$ are used from the set $\left\{0.2, 0.4, 0.6, 0.8, 1.0\right\}$. We can find that the variation of $\alpha$ is not very obvious for the improvement of the transferability, which is because our attack method uses an iterative approach, $\alpha$ can be likened to the step size, and iteration can be likened to the number of steps. And under the influence of the number of steps, the small variations of $\alpha$ will not have any effect on the attainable destination, i.e., generating transferable adversarial point clouds.


\paragraph{Effect of Feature Layer $k$ on Attack Success Rates}
As shown in \figurename~\ref{fig:6}, we further analyze the effectiveness of our method's choice of feature layer $k$. Here, DGCNN is chosen as the source model to generate adversarial point clouds to attack other target models, as a way to explore the impact of the choice of feature layer $k$ on the transferability. Obviously, we can see that the success rate of the attack on the target model is highest when choosing the feature layer Conv2. This is because the features extracted from different layers of the DNNs can be quite different from each other. For Conv1, the model has not yet learned the critical features that are important for the classification, whereas for Conv3 and Conv4, after multi-layer learning, at this point, the features tend to be specific to the source model, and the problem of overfitting to the source model in the white-box attack occurs. Overall, the features in Conv2 are the most appropriate, where this layer can extract critical features without overfitting the source model.

\section{Conclusion}

In this paper, we propose a novel 3D transfer-based black-box attack that introduces critical features as a guide to make the search of the adversarial point clouds toward the shared vulnerable area among different models, thus significantly improving the transferability of the adversarial point clouds. We construct a new loss function and combine an iterative search to find the optimal adversarial point cloud. In future work, we would extend this attack to point cloud segmentation. As segmentation models rely on local structures, we would introduce a graph convolutional network to ensure that attacked points maintain spatial consistency.

\section*{Acknowledgement}
This work is supported by the Yunnan Province Science and Technology Major Project (Grant No.202302AD080002), National Natural Science Foundation of China (Grant No.62206128), National Key Research and Development Program of China under (Grant No.2023YFB2703900) and the Start-up Fund for New Talented Researchers of Nanjing University of Industry Technology (Grant No. YK24-05-04).
Liming Lu is supported by the Postgraduate Research \& Practice Innovation Program of Jiangsu Province (Grant No.KYCX24\_0723).
{
    \small
    \bibliographystyle{unsrt}
    \bibliography{Ref}

\begin{thebibliography}{10}

\bibitem{lu2024uniads}
Liming Lu, Zhenghan Chen, Xiaoyu Lu, Yihang Rao, Lujun Li, and Shuchao Pang.
\newblock Uniads: Universal architecture-distiller search for distillation gap.
\newblock In {\em Proceedings of the AAAI Conference on Artificial Intelligence}, volume~38, pages 14167--14174, 2024.

\bibitem{xu2024enhancing}
Wentao Xu, Qianqian Xie, Shuo Yang, Jiangxia Cao, and Shuchao Pang.
\newblock Enhancing content-based recommendation via large language model.
\newblock In {\em Proceedings of the 33rd ACM International Conference on Information and Knowledge Management}, pages 4153--4157, 2024.

\bibitem{pang2025pridm}
Shuchao Pang, Yihang Rao, Zhigang Lu, Haichen Wang, Yongbin Zhou, and Minhui Xue.
\newblock Pridm: Effective and universal private data recovery via diffusion models.
\newblock {\em IEEE Transactions on Dependable and Secure Computing}, 2025.

\bibitem{wang2024dp}
Haichen Wang, Shuchao Pang, Zhigang Lu, Yihang Rao, Yongbin Zhou, and Minhui Xue.
\newblock dp-promise: Differentially private diffusion probabilistic models for image synthesis.
\newblock In {\em 33rd USENIX Security Symposium (USENIX Security 24)}, pages 1063--1080, 2024.

\bibitem{xu2024memory}
Xiuwei Xu, Chong Xia, Ziwei Wang, Linqing Zhao, Yueqi Duan, Jie Zhou, and Jiwen Lu.
\newblock Memory-based adapters for online 3d scene perception.
\newblock In {\em Proceedings of the IEEE/CVF Conference on Computer Vision and Pattern Recognition}, pages 21604--21613, 2024.

\bibitem{jiang2023symphonize}
Haoyi Jiang, Tianheng Cheng, Naiyu Gao, Haoyang Zhang, Tianwei Lin, Wenyu Liu, and Xinggang Wang.
\newblock Symphonize 3d semantic scene completion with contextual instance queries.
\newblock In {\em Proceedings of the IEEE/CVF Conference on Computer Vision and Pattern Recognition}, pages 20258--20267, 2024.

\bibitem{huang2023ptt}
Kuan-Chih Huang, Weijie Lyu, Ming-Hsuan Yang, and Yi-Hsuan Tsai.
\newblock Ptt: Point-trajectory transformer for efficient temporal 3d object detection.
\newblock In {\em Proceedings of the IEEE/CVF Conference on Computer Vision and Pattern Recognition}, pages 14938--14947, 2024.

\bibitem{liu2024hqa}
Han Liu, Zhi Xu, Xiaotong Zhang, Feng Zhang, Fenglong Ma, Hongyang Chen, Hong Yu, and Xianchao Zhang.
\newblock Hqa-attack: Toward high quality black-box hard-label adversarial attack on text.
\newblock {\em Advances in Neural Information Processing Systems}, 36, 2024.

\bibitem{wu2024towards}
Shangbo Wu, Yu-an Tan, Yajie Wang, Ruinan Ma, Wencong Ma, and Yuanzhang Li.
\newblock Towards transferable adversarial attacks with centralized perturbation.
\newblock In {\em Proceedings of the AAAI Conference on Artificial Intelligence}, volume~38, pages 6109--6116, 2024.

\bibitem{liu2024explicitly}
Daizong Liu and Wei Hu.
\newblock Explicitly perceiving and preserving the local geometric structures for 3d point cloud attack.
\newblock In {\em Proceedings of the AAAI Conference on Artificial Intelligence}, volume~38, pages 3576--3584, 2024.

\bibitem{65}
Chalavadi Vishnu, Jayesh Khandelwal, C~Krishna Mohan, and Cenkeramaddi~Linga Reddy.
\newblock Ev aa-exchange vanishing adversarial attack on lidar point clouds in autonomous vehicles.
\newblock {\em IEEE Transactions on Geoscience and Remote Sensing}, 61:5703410, 2023.

\bibitem{cheng2023topology}
Riran Cheng, Xupeng Wang, Ferdous Sohel, and Hang Lei.
\newblock Topology-aware universal adversarial attack on 3d object tracking.
\newblock {\em Visual Intelligence}, 1(1):31, 2023.

\bibitem{kong2024patch}
Dehong Kong, Siyuan Liang, Xiaopeng Zhu, Yuansheng Zhong, and Wenqi Ren.
\newblock Patch is enough: naturalistic adversarial patch against vision-language pre-training models.
\newblock {\em Visual Intelligence}, 2(1):33, 2024.

\bibitem{liang2021generate}
Siyuan Liang, Xingxing Wei, and Xiaochun Cao.
\newblock Generate more imperceptible adversarial examples for object detection.
\newblock In {\em ICML 2021 Workshop on Adversarial Machine Learning}, 2021.

\bibitem{liang2020efficient}
Siyuan Liang, Xingxing Wei, Siyuan Yao, and Xiaochun Cao.
\newblock Efficient adversarial attacks for visual object tracking.
\newblock In {\em Computer Vision--ECCV 2020: 16th European Conference, Glasgow, UK, August 23--28, 2020, Proceedings, Part XXVI 16}, 2020.

\bibitem{wei2018transferable}
Xingxing Wei, Siyuan Liang, Ning Chen, and Xiaochun Cao.
\newblock Transferable adversarial attacks for image and video object detection.
\newblock {\em arXiv preprint arXiv:1811.12641}, 2018.

\bibitem{liang2022parallel}
Siyuan Liang, Baoyuan Wu, Yanbo Fan, Xingxing Wei, and Xiaochun Cao.
\newblock Parallel rectangle flip attack: A query-based black-box attack against object detection.
\newblock {\em arXiv preprint arXiv:2201.08970}, 2022.

\bibitem{liang2022large}
Siyuan Liang, Longkang Li, Yanbo Fan, Xiaojun Jia, Jingzhi Li, Baoyuan Wu, and Xiaochun Cao.
\newblock A large-scale multiple-objective method for black-box attack against object detection.
\newblock In {\em European Conference on Computer Vision}, 2022.

\bibitem{xiang2019generating}
Chong Xiang, Charles~R Qi, and Bo~Li.
\newblock Generating 3d adversarial point clouds.
\newblock In {\em Proceedings of the IEEE/CVF Conference on Computer Vision and Pattern Recognition}, pages 9136--9144, 2019.

\bibitem{wen2020geometry}
Yuxin Wen, Jiehong Lin, Ke~Chen, CL~Philip Chen, and Kui Jia.
\newblock Geometry-aware generation of adversarial point clouds.
\newblock {\em IEEE Transactions on Pattern Analysis and Machine Intelligence}, 44(6):2984--2999, 2020.

\bibitem{60}
Qidong Huang, Xiaoyi Dong, Dongdong Chen, Hang Zhou, Weiming Zhang, and Nenghai Yu.
\newblock Shape-invariant 3d adversarial point clouds.
\newblock In {\em Proceedings of the IEEE/CVF Conference on Computer Vision and Pattern Recognition}, pages 15335--15344, 2022.

\bibitem{tao20233dhacker}
Yunbo Tao, Daizong Liu, Pan Zhou, Yulai Xie, Wei Du, and Wei Hu.
\newblock 3dhacker: Spectrum-based decision boundary generation for hard-label 3d point cloud attack.
\newblock In {\em Proceedings of the IEEE/CVF International Conference on Computer Vision}, pages 14340--14350, 2023.

\bibitem{hamdi2020advpc}
Abdullah Hamdi, Sara Rojas, Ali Thabet, and Bernard Ghanem.
\newblock Advpc: Transferable adversarial perturbations on 3d point clouds.
\newblock In {\em European Conference on Computer Vision}, pages 241--257, 2020.

\bibitem{he2023generating}
Bangyan He, Jian Liu, Yiming Li, Siyuan Liang, Jingzhi Li, Xiaojun Jia, and Xiaochun Cao.
\newblock Generating transferable 3d adversarial point cloud via random perturbation factorization.
\newblock In {\em Proceedings of the AAAI Conference on Artificial Intelligence}, volume~37, pages 764--772, 2023.

\bibitem{ganeshan2019fda}
Aditya Ganeshan, Vivek BS, and R~Venkatesh Babu.
\newblock Fda: Feature disruptive attack.
\newblock In {\em Proceedings of the IEEE/CVF International Conference on Computer Vision}, pages 8069--8079, 2019.

\bibitem{wang2021feature}
Zhibo Wang, Hengchang Guo, Zhifei Zhang, Wenxin Liu, Zhan Qin, and Kui Ren.
\newblock Feature importance-aware transferable adversarial attacks.
\newblock In {\em Proceedings of the IEEE/CVF International Conference on Computer Vision}, pages 7639--7648, 2021.

\bibitem{wu2020boosting}
Weibin Wu, Yuxin Su, Xixian Chen, Shenglin Zhao, Irwin King, Michael~R Lyu, and Yu-Wing Tai.
\newblock Boosting the transferability of adversarial samples via attention.
\newblock In {\em Proceedings of the IEEE/CVF Conference on Computer Vision and Pattern Recognition}, pages 1161--1170, 2020.

\bibitem{67}
Wu~Zheng, Mingxuan Hong, Li~Jiang, and Chi-Wing Fu.
\newblock Boosting 3d object detection by simulating multimodality on point clouds.
\newblock In {\em Proceedings of the IEEE/CVF Conference on Computer Vision and Pattern Recognition}, pages 13638--13647, 2022.

\bibitem{68}
Mengtian Li, Yuan Xie, Yunhang Shen, Bo~Ke, Ruizhi Qiao, Bo~Ren, Shaohui Lin, and Lizhuang Ma.
\newblock Hybridcr: Weakly-supervised 3d point cloud semantic segmentation via hybrid contrastive regularization.
\newblock In {\em Proceedings of the IEEE/CVF Conference on Computer Vision and Pattern Recognition}, pages 14930--14939, 2022.

\bibitem{69}
Shitong Luo, Jiahan Li, Jiaqi Guan, Yufeng Su, Chaoran Cheng, Jian Peng, and Jianzhu Ma.
\newblock Equivariant point cloud analysis via learning orientations for message passing.
\newblock In {\em Proceedings of the IEEE/CVF Conference on Computer Vision and Pattern Recognition}, pages 18932--18941, 2022.

\bibitem{70}
Zhimin Yuan, Ming Cheng, Wankang Zeng, Yanfei Su, Weiquan Liu, Shangshu Yu, and Cheng Wang.
\newblock Prototype-guided multitask adversarial network for cross-domain lidar point clouds semantic segmentation.
\newblock {\em IEEE Transactions on Geoscience and Remote Sensing}, 61:570013, 2023.

\bibitem{qi2017pointnet}
Charles~R Qi, Hao Su, Kaichun Mo, and Leonidas~J Guibas.
\newblock Pointnet: Deep learning on point sets for 3d classification and segmentation.
\newblock In {\em Proceedings of the IEEE Conference on Computer Vision and Pattern Recognition}, pages 652--660, 2017.

\bibitem{qi2017pointnet++}
Charles~Ruizhongtai Qi, Li~Yi, Hao Su, and Leonidas~J Guibas.
\newblock Pointnet++: Deep hierarchical feature learning on point sets in a metric space.
\newblock {\em Advances in Neural Information Processing Systems}, pages 5099--5108, 2017.

\bibitem{wang2019dynamic}
Yue Wang, Yongbin Sun, Ziwei Liu, Sanjay~E Sarma, Michael~M Bronstein, and Justin~M Solomon.
\newblock Dynamic graph cnn for learning on point clouds.
\newblock {\em ACM Transactions on Graphics}, 38(5):1--12, 2019.

\bibitem{wu2019pointconv}
Wenxuan Wu, Zhongang Qi, and Li~Fuxin.
\newblock Pointconv: Deep convolutional networks on 3d point clouds.
\newblock In {\em Proceedings of the IEEE/CVF Conference on Computer Vision and Pattern Recognition}, pages 9621--9630, 2019.

\bibitem{xiang2021walk}
Tiange Xiang, Chaoyi Zhang, Yang Song, Jianhui Yu, and Weidong Cai.
\newblock Walk in the cloud: Learning curves for point clouds shape analysis.
\newblock In {\em Proceedings of the IEEE/CVF International Conference on Computer Vision}, pages 915--924, 2021.

\bibitem{53}
Meng-Hao Guo, Jun-Xiong Cai, Zheng-Ning Liu, Tai-Jiang Mu, Ralph~R Martin, and Shi-Min Hu.
\newblock Pct: Point cloud transformer.
\newblock {\em Computational Visual Media}, 7:187--199, 2021.

\bibitem{59}
Renrui Zhang, Liuhui Wang, Yali Wang, Peng Gao, Hongsheng Li, and Jianbo Shi.
\newblock Parameter is not all you need: Starting from non-parametric networks for 3d point cloud analysis.
\newblock {\em arXiv preprint arXiv:2303.08134}, 2023.

\bibitem{13}
Daniel Liu, Ronald Yu, and Hao Su.
\newblock Extending adversarial attacks and defenses to deep 3d point cloud classifiers.
\newblock In {\em International Conference on Image Processing}, pages 2279--2283, 2019.

\bibitem{tsai2020robust}
Tzungyu Tsai, Kaichen Yang, Tsung-Yi Ho, and Yier Jin.
\newblock Robust adversarial objects against deep learning models.
\newblock In {\em Proceedings of the AAAI Conference on Artificial Intelligence}, volume~34, pages 954--962, 2020.

\bibitem{liu2022boosting}
Binbin Liu, Jinlai Zhang, and Jihong Zhu.
\newblock Boosting 3d adversarial attacks with attacking on frequency.
\newblock {\em IEEE Access}, 10:50974--50984, 2022.

\bibitem{34}
Jiancheng Yang, Qiang Zhang, Rongyao Fang, Bingbing Ni, Jinxian Liu, and Qi~Tian.
\newblock Adversarial attack and defense on point sets.
\newblock {\em arXiv preprint arXiv:1902.10899}, 2019.

\bibitem{16}
Hang Zhou, Kejiang Chen, Weiming Zhang, Han Fang, Wenbo Zhou, and Nenghai Yu.
\newblock Dup-net: Denoiser and upsampler network for 3d adversarial point clouds defense.
\newblock In {\em Proceedings of the IEEE/CVF International Conference on Computer Vision}, pages 1961--1970, 2019.

\bibitem{22}
Ziyi Wu, Yueqi Duan, He~Wang, Qingnan Fan, and Leonidas~J Guibas.
\newblock If-defense: 3d adversarial point cloud defense via implicit function based restoration.
\newblock {\em arXiv preprint arXiv:2010.05272}, 2020.

\bibitem{39}
Guanlin Li, Guowen Xu, Han Qiu, Ruan He, Jiwei Li, and Tianwei Zhang.
\newblock Improving adversarial robustness of 3d point cloud classification models.
\newblock In {\em European Conference on Computer Vision}, pages 672--689, 2022.

\bibitem{21}
Jinlai Zhang, Lyujie Chen, Bo~Ouyang, Binbin Liu, Jihong Zhu, Yujin Chen, Yanmei Meng, and Danfeng Wu.
\newblock Pointcutmix: Regularization strategy for point cloud classification.
\newblock {\em Neurocomputing}, 505:58--67, 2022.

\bibitem{zheng2019pointcloud}
Tianhang Zheng, Changyou Chen, Junsong Yuan, Bo~Li, and Kui Ren.
\newblock Pointcloud saliency maps.
\newblock In {\em Proceedings of the IEEE/CVF International Conference on Computer Vision}, pages 1598--1606, 2019.

\bibitem{wu20153d}
Zhirong Wu, Shuran Song, Aditya Khosla, Fisher Yu, Linguang Zhang, Xiaoou Tang, and Jianxiong Xiao.
\newblock 3d shapenets: A deep representation for volumetric shapes.
\newblock In {\em Proceedings of the IEEE Conference on Computer Vision and Pattern Recognition}, pages 1912--1920, 2015.

\bibitem{uy2019revisiting}
Mikaela~Angelina Uy, Quang-Hieu Pham, Binh-Son Hua, Thanh Nguyen, and Sai-Kit Yeung.
\newblock Revisiting point cloud classification: A new benchmark dataset and classification model on real-world data.
\newblock In {\em Proceedings of the IEEE/CVF International Conference on Computer Vision}, pages 1588--1597, 2019.

\bibitem{4}
R~Qi Charles, Hao Su, Mo~Kaichun, and Leonidas~J Guibas.
\newblock Pointnet: Deep learning on point sets for 3d classification and segmentation.
\newblock In {\em Proceedings of the IEEE/CVF Conference on Computer Vision and Pattern Recognition}, pages 77--85, 2017.

\bibitem{liu2022imperceptible}
Daizong Liu and Wei Hu.
\newblock Imperceptible transfer attack and defense on 3d point cloud classification.
\newblock {\em IEEE Transactions on Pattern Analysis and Machine Intelligence}, 45(4):4727--4746, 2022.

\bibitem{li2018pointcnn}
Yangyan Li, Rui Bu, Mingchao Sun, Wei Wu, Xinhan Di, and Baoquan Chen.
\newblock Pointcnn: Convolution on x-transformed points.
\newblock {\em Advances in Neural Information Processing Systems}, pages 820--830, 2018.

\bibitem{zhao2021point}
Hengshuang Zhao, Li~Jiang, Jiaya Jia, Philip~HS Torr, and Vladlen Koltun.
\newblock Point transformer.
\newblock In {\em Proceedings of the IEEE/CVF International Conference on Computer Vision}, pages 16259--16268, 2021.

\bibitem{liu2023exploring}
Aishan Liu, Shiyu Tang, Siyuan Liang, Ruihao Gong, Boxi Wu, Xianglong Liu, and Dacheng Tao.
\newblock Exploring the relationship between architectural design and adversarially robust generalization.
\newblock In {\em Proceedings of the IEEE/CVF Conference on Computer Vision and Pattern Recognition}, 2023.

\bibitem{lu2025adversarial}
Liming Lu, Shuchao Pang, Siyuan Liang, Haotian Zhu, Xiyu Zeng, Aishan Liu, Yunhuai Liu, and Yongbin Zhou.
\newblock Adversarial training for multimodal large language models against jailbreak attacks.
\newblock {\em arXiv preprint arXiv:2503.04833}, 2025.

\bibitem{zhang2024lanevil}
Tianyuan Zhang, Lu~Wang, Hainan Li, Yisong Xiao, Siyuan Liang, Aishan Liu, Xianglong Liu, and Dacheng Tao.
\newblock Lanevil: Benchmarking the robustness of lane detection to environmental illusions.
\newblock {\em arXiv preprint arXiv:2406.00934}, 2024.

\bibitem{liang2023exploring}
Jiawei Liang, Siyuan Liang, Aishan Liu, Ke~Ma, Jingzhi Li, and Xiaochun Cao.
\newblock Exploring inconsistent knowledge distillation for object detection with data augmentation.
\newblock In {\em Proceedings of the 31st ACM International Conference on Multimedia}, 2023.

\end{thebibliography}
}

\end{document}